\documentclass[journal,twoside,web]{ieeecolor}
\usepackage{generic}
\usepackage{cite}
\usepackage{amsmath,amssymb,amsfonts}
\usepackage{algorithmic}
\usepackage{graphicx}
\usepackage{algorithm,algorithmic}
\usepackage{hyperref}
\usepackage{textcomp}
\usepackage{multirow}

\def\BibTeX{{\rm B\kern-.05em{\sc i\kern-.025em b}\kern-.08em
    T\kern-.1667em\lower.7ex\hbox{E}\kern-.125emX}}

\begin{document}
    
\markboth{\hskip25pc IEEE Journal of Biomedical and Health Informatics}
         {P. Wang \MakeLowercase{\textit{et al.}}: ConDistFL—Conditional Distillation for FL}

\title{Conditional Distillation for Federated Learning with Partially Labeled Data}

\author{
Pochuan Wang,
Chen Shen,
Masahiro Oda,
Chiou-Shann Fuh,
Kensaku Mori,
Weichung Wang,
and Holger R. Roth
\thanks{Manuscript submitted August 7, 2025, to \textit{IEEE Journal of Biomedical and Health Informatics}.  
This work was supported in part by the National Science and Technology Council (NSTC), Taiwan, under Grants NSTC 112-2634-F-002-003- and NSTC 113-2123-M-002-017-, and in part by the Ministry of Health and Welfare, Taiwan, under Grant MOHW113-TDU-B221-134003.  
The authors also thank the National Center for High-performance Computing (NCHC) for providing computational and storage resources. The kidney, liver, pancreas, and spleen icons in Figure~\ref{fig:fl_setup} were created using resources from Flaticon.com.}
\thanks{P. Wang and C.-S. Fuh are with the Department of Computer Science and Information Engineering, National Taiwan University, Taipei 10617, Taiwan (e-mail: \{d08922016, fuh\}@csie.ntu.edu.tw).}
\thanks{C. Shen, M. Oda, and K. Mori are with the Graduate School of Informatics, Nagoya University, Nagoya 464-8601, Japan (e-mail: kensaku@is.nagoya-u.ac.jp, \{cshen, moda\}@mori.m.is.nagoya-u.ac.jp).}
\thanks{W. Wang is with the Institute of Applied Mathematical Sciences, National Taiwan University, Taipei 10617, Taiwan (e-mail: wwang@ntu.edu.tw).}
\thanks{H. R. Roth is with NVIDIA Corporation, Santa Clara, CA 20814 USA (e-mail: hroth@nvidia.com).}
\thanks{Corresponding authors: Kensaku Mori (kensaku@is.nagoya-u.ac.jp), Weichung Wang (wwang@ntu.edu.tw), and Holger R. Roth (hroth@nvidia.com).}
}

\maketitle

\begin{abstract}

Multi-organ segmentation underpins diagnosis and treatment planning, yet fully annotated, shareable datasets are scarce due to cost and privacy constraints. Federated learning (FL) enables collaboration without sharing data, but inconsistent client labels cause model divergence and catastrophic forgetting. We propose ConDistFL, a label-aware FL framework that uses conditional distillation to leverage partially labeled data while avoiding contradictory supervision. At each FL round, the aggregated global model acts as a teacher. Clients compute a conditional distillation loss for locally unlabeled classes from the discrepancy between global and local predictions. This enables stable knowledge transfer across clients with non-overlapping label sets. We evaluate ConDistFL in realistic federated settings on 3D abdominal computed tomography (CT) and 2D chest X-ray (CXR). Each client contributes labels for only one structure. ConDistFL attains the highest mean Dice among federated baselines on both modalities. It mitigates catastrophic forgetting caused by locally missing annotations for specific classes and improves local performance on those classes. The model generalizes to the external AMOS22 cohort with unseen contrast phases, indicating resilience to domain shift. ConDistFL is efficient and scalable: conditional distillation adds only minor client-side computation, and the communication volume per round is unchanged from standard FedAvg. Its architecture-agnostic design integrates readily with existing FL workflows and model architectures, making the method practical for multi-center deployment. By addressing partial labeling while preserving data privacy, ConDistFL advances collaborative medical image analysis and enables multi-institutional studies.
\end{abstract}

\begin{IEEEkeywords}
Computed tomography,
Deep learning,
Federated learning,
Knowledge distillation,
Medical image segmentation,
Privacy,
Tumors,
X-ray imaging
\end{IEEEkeywords}

\section{Introduction}
\label{sec:introduction}

Accurate segmentation of multiple organs and lesions plays a critical role in medical imaging, supporting a wide range of clinical applications such as disease diagnosis, treatment planning, and surgical guidance. With the growing demand for comprehensive AI-based image analysis tools, robust and generalizable multi-organ segmentation models are increasingly essential. However, obtaining fully annotated datasets for training these models is extremely costly and labor-intensive, often requiring extensive input from expert radiologists. Additionally, strict privacy regulations further complicate data sharing across institutions, limiting the ability to pool data for centralized training.

While several public datasets provide multi-organ segmentations, such as Beyond the Cranial Vault (BTCV)~\cite{landman2015miccai}, AMOS22~\cite{ji2022amos}, and TotalSegmentator~\cite{Wasserthal2023-ry}, most focus exclusively on healthy anatomical structures and lack lesion or tumor annotations. In contrast, datasets that include lesion segmentation, such as the Medical Segmentation Decathlon (MSD)~\cite{Antonelli2021-io} and KiTS19~\cite{Heller2021}, are often limited to single-organ tasks and lack broader coverage. This scarcity of tumor-annotated data poses a major barrier to developing models that can accurately segment clinically significant abnormalities across multiple organs.

\begin{figure}[ht]
  \centering
  \includegraphics[width=\columnwidth]{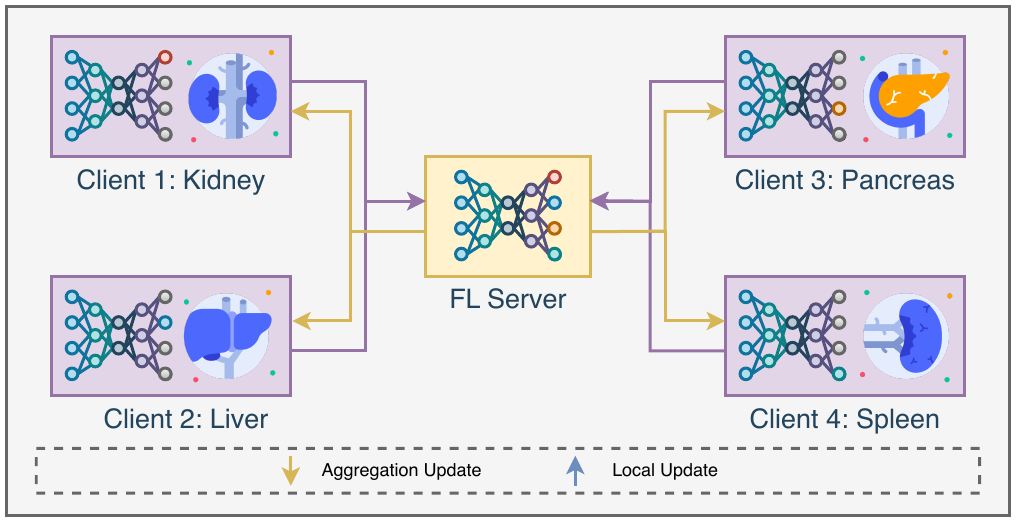}
  \caption{An illustration of the federated learning setup for multi-organ and tumor segmentation using inconsistently labeled datasets. Each client contributes annotations for only a subset of target organs or lesions, reflecting realistic data availability in multi-institutional settings.}
  \label{fig:fl_setup}
\end{figure}

Federated Learning (FL) has emerged as a promising solution to address privacy concerns by enabling decentralized training across institutions without requiring direct data sharing~\cite{rieke2020future}. However, FL encounters significant challenges in real-world medical applications where datasets are inconsistently labeled across sites. In such settings, each participating client may have annotations for only a subset of organs or lesions, leading to partially and inconsistently labeled data distributions. This setup is illustrated in Figure \ref{fig:fl_setup}, where each client contributes segmentation labels for only a single organ (and its associated lesion, if available), reflecting realistic cross-institutional annotation gaps in federated learning scenarios. This partial labeling introduces two critical challenges for FL: (1) model divergence, where updates from clients with heterogeneous data distributions lead to degraded global model performance due to conflicting gradient directions, and (2) catastrophic forgetting, where knowledge about previously seen structures is lost over subsequent training rounds due to incomplete label coverage.

To address these challenges, we propose \textbf{ConDistFL}, a novel federated learning framework that integrates conditional distillation into the FL process to effectively leverage inconsistently labeled data. By decoupling the supervised learning on local partially labeled data and the distillation-based learning from the global model's predictions, ConDistFL mitigates both model divergence and catastrophic forgetting. This design enables knowledge transfer across clients even in the absence of overlapping labels, thereby improving robustness and generalizability.

The key contributions of this work include:
\begin{itemize}
    \item \textbf{Conditional Distillation Loss:} A label-aware distillation loss reducing supervision conflicts, model divergence, and catastrophic forgetting.
    
    \item \textbf{Tumor Segmentation in Federated Settings:} Accurate lesion segmentation without fully annotated clients, surpassing existing federated methods.
    
    \item \textbf{Robust Out-of-Federation Generalization:} Improved Dice scores on external datasets with unseen contrast phases, demonstrating strong domain robustness.
    
    \item \textbf{Cross-Modality Robustness:} Effective generalization to both 3D abdominal CT and 2D chest X-ray, indicating broad clinical applicability.
    
    \item \textbf{Efficiency and Scalability:} Efficient computation and communication, suitable for practical multi-center clinical deployments.
\end{itemize}

This work extends our earlier workshop paper~\cite{condistFL} by including evaluations on 2D chest X-ray data, a comprehensive ablation study, Dice score distribution analysis of local models, and in-depth analysis of the ConDist loss. These additions provide a substantially broader and more comprehensive validation of the proposed approach.

\section{Related Works}
\label{sec:related_works}

In this section, we review the key advancements relevant to our study, focusing on three areas: segmentation models, centralized learning for partially labeled data, and FL approaches for inconsistent label scenarios.

\subsection{Centralized Methods for Medical Image Segmentation}
\label{sec:related_centralized}

Deep learning has significantly advanced medical image segmentation, with U-Net-based models~\cite{ronneberger2015u, vnet2016, Zhou2018-ji} widely used in clinical tasks. nnU-Net~\cite{isensee2021nnu} notably introduced automated, dataset-specific model adaptation, establishing a robust benchmark for research and deployment. Recent CNN-transformer hybrids, such as UNetR~\cite{hatamizadeh2022unetr} and SwinUNetR~\cite{hatamizadeh2021swin}, improved global context representation and computational efficiency. Among recent architectures, MedNeXt~\cite{roy2023mednext}, built from scalable 3D ConvNeXt blocks, has demonstrated state-of-the-art results on multiple public datasets~\cite{Isensee2024-fr}. Owing to its strong accuracy for volumetric data, we adopt MedNeXt as the backbone for all experiments in this work.

Although most deep segmentation models assume access to complete, uniformly labeled datasets, this assumption is rarely met in multi-center studies or when working with real clinical data. In practice, datasets often differ in their annotated structures, motivating the development of methods that can leverage partially labeled data without discarding valuable supervision.

Several centralized approaches have been proposed for partially labeled datasets. Some adapt their loss functions explicitly to manage missing annotations, such as marginal and exclusive losses~\cite{SHI2021101979}. Architectures like PIPO-FAN~\cite{fang2020multi} dynamically fuse supervision across heterogeneous sources using multi-scale and target-adaptive designs. Task-conditioned models employ organ-specific queries to segment varying structures~\cite{zhang2021dodnet, xie2023learning}. Other methods use pseudo-labeling for iterative self-training~\cite{liu2024cosst} or CLIP-based language–image embeddings for sparse annotation scenarios~\cite{liu2023clip}. Incremental learning frameworks~\cite{Liu2022-ec, Ji2023-vw} progressively incorporate new datasets but require repeated centralized data access.

Although these centralized methods effectively handle partial labels, their dependence on pooling patient data across institutions conflicts with patient privacy and regulatory requirements, highlighting the need for FL solutions that preserve data sovereignty while enabling label-efficient training.

\subsection{Federated Learning for Partially Labeled Datasets}

FL methods address privacy concerns by training models across institutions without data sharing. However, they face unique challenges when labels are inconsistently distributed. In such settings, models suffer from \textbf{model divergence} due to non-i.i.d. data distributions, and \textbf{catastrophic forgetting} of classes not present on a given client.

Our earlier work~\cite{shen2022joint} introduced sigmoid activations to mitigate cross-client label conflicts; however, this approach neither addressed model divergence nor prevented catastrophic forgetting, resulting in the global model's inability to recognize all classes. MenuNet~\cite{xu2023federated} reduces forgetting by partitioning the feature extractor into organ-specific encoders, freezing parameters on clients lacking corresponding annotations, and alleviates divergence by including an additional fully annotated client. FedIOD~\cite{wan2024fediod} addresses divergence by modeling inter-organ dependencies through transformer blocks placed between a shared encoder and organ-specific decoders; similar to MenuNet, it avoids catastrophic forgetting by freezing decoder parameters for organs absent on a given client. Kim et al.~\cite{Kim2024-ci} adopted a knowledge distillation (KD) strategy, where each client infers soft predictions from a global model and integrates auxiliary predictions generated locally using peer models independently pre-trained at other sites, thus mitigating both divergence and forgetting.

These approaches collectively exhibit practical limitations. Parameter duplication in methods using separate encoders or decoders inflates GPU memory and computational requirements. Solutions relying on spatial correspondence or anatomical regularity struggle when lesions or structures deviate from expected positions. Finally, distillation methods involving multiple teacher models require additional computational costs and may introduce inconsistencies when combining unaligned soft predictions for teachers and local ground-truth labels.

ConDistFL enhances the knowledge distillation framework by introducing a label-aware conditional distillation loss. Each client, upon receiving the global model, learns from both local hard labels and the global model's soft predictions without necessitating additional teacher networks or pre-training. The conditional loss masks voxels with reliable ground truth annotations, thereby avoiding contradictory supervision and facilitating smooth convergence. This innovation, embedded within the loss function, is architecture-agnostic, does not increase parameter count, and scales efficiently with the number of organs. Its straightforward integration with 2D and 3D segmentation models and robust generalization across various modalities and external datasets underscores ConDistFL's practicality for multi-center deployment.

\section{Method}
\label{sec:method}

We introduce ConDistFL in detail. The primary challenge we address is the partial labeling across different clients, where each client's dataset contains labels for only a subset of the total organ classes. The proposed method integrates a novel Conditional Distillation (ConDist) loss with traditional supervised learning to enhance the performance and generalizability of segmentation models in a federated setting. Our framework, therefore, optimizes a combined loss consisting of a supervised loss and the ConDist loss (detailed in~\ref{sec:condist_loss}). The implementation of ConDistFL is publicly available at \url{https://github.com/NVIDIA/NVFlare/tree/main/research/condist-fl}.

Let $\mathcal{F}$ represent the segmentation model, which aims to identify $N$ classes of regions of interest (ROIs) across $K$ clients, including the background. For an image $x_{k}$ in client $k \in \{1,\dots,K\}$, only a subset of these $N$ classes are annotated by radiologists in its corresponding label $y_{k}$, denoted as the foreground classes $F_{k}$ for client $k$. The set of remaining classes, including the background class $0$ and all unlabeled classes for client $k$, is denoted as $B_{k}$. The union of all foreground classes across clients must collectively cover all $N-1$ non-background classes, ensuring that $|F_{1} \cup F_{2} \cup \dots \cup F_{K}| = N-1$.

\subsection{Output Activation and Softmax Function}

In multi-organ segmentation, each pixel in an image must be assigned exclusively to one of the $N$ classes, as human organs do not overlap. To enforce this constraint, we adopt the softmax function $\sigma$ (Equation~\ref{eq:softmax}) as the output activation. This function transforms the output logits into a probability distribution over the $N$ classes, ensuring that the model's output reflects the mutually exclusive nature of organ segmentation.

\begin{equation}
  \label{eq:softmax}
  \sigma_{i}(z) = \frac{e^{z_i}}{\sum^{N}_{j=1}{e^{z_j}}}
\end{equation}

The use of softmax differs from a sigmoid-based activation, where each class would be treated independently. The softmax formulation prevents overlapping segmentations and encourages the model to learn spatial relationships between different organs.

\subsection{Training Process and Loss Functions}

\begin{figure*}[htb]
  \centering
  \includegraphics[width=0.8\textwidth]{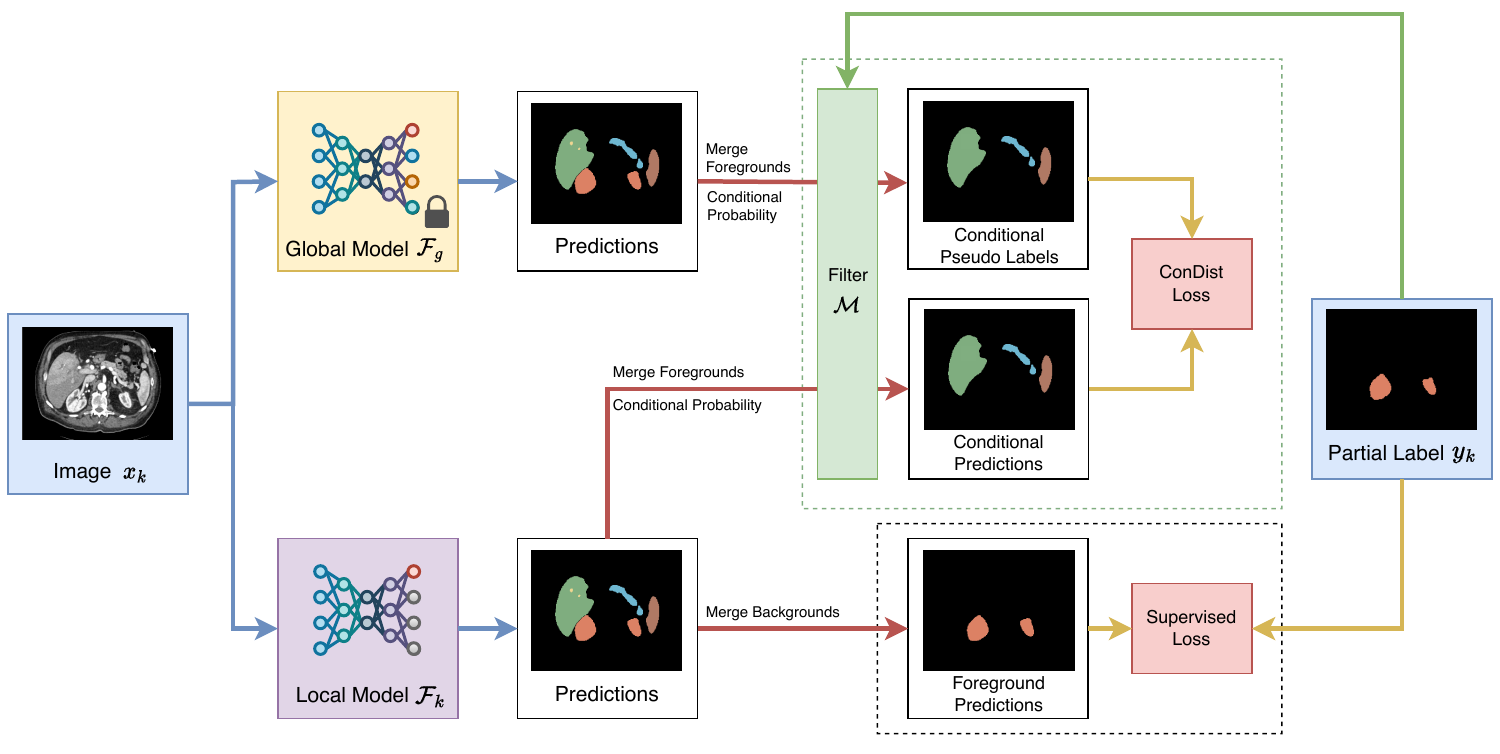}
  \caption{Illustration of the ConDistFL client training process on a partially labeled dataset, where only the kidney and kidney tumor are annotated. The model is trained to leverage both local and global knowledge for effective segmentation of labeled and unlabeled organs.}
  \label{fig:condist_client}
\end{figure*}

The ConDistFL training process, as illustrated in Figure \ref{fig:condist_client}, involves two key components: a supervised learning loss $\mathcal{L}_{sup}$ (details in Section \ref{sec:sup_loss}) and a ConDist loss $\mathcal{L}_{\text{ConDist}}$ (details in Section \ref{sec:condist_loss}). The supervised loss handles the labeled data available at each client, while the ConDist loss leverages the global model’s predictions to enhance the local model’s performance on unlabeled classes.

\subsection{Supervised Learning Loss}
\label{sec:sup_loss}

One key challenge in FL with partially labeled data is that unlabeled classes in $B_{k}$ are incorrectly assigned to the background class. This erroneous assignment introduces inconsistencies in target definitions across clients, negatively impacting FL convergence. To mitigate this issue, we adopt the marginal loss~\cite{SHI2021101979}, which adjusts the probability distributions of both the ground truth labels $y_{k}$ and the model’s predicted output $\hat{y}_{k}$. Here, $y_{k}$ represents the one-hot encoded ground truth, while $\hat{y}_{k} = \sigma(\mathcal{F}_{k}(x_{k}))$ is the softmax-normalized output of the local model on client $k$.

The marginal loss approach preserves the probability distribution of the foreground classes $F_{k}$ while accumulating the probability mass for the background and the unlabeled classes in $B_{k}$. This ensures unlabeled classes are not mistakenly assigned to the background category, resolving the label conflict across clients. We implement the supervised loss $\mathcal{L}_{sup}$ by combining marginal Dice loss with cross-entropy loss, optimizing the model for segmentation performance.

\subsection{Conditional Distillation Loss}
\label{sec:condist_loss}

The ConDist loss $\mathcal{L}_{\text{ConDist}}$ minimizes the discrepancy between the global model’s predictions and the local model’s predictions for unlabeled classes. This loss resembles knowledge distillation, employing the softmax function with a temperature parameter $\tau$ to smooth output probability distributions. Let $\hat{y}_{g}^{\tau}=\sigma(\mathcal{F}_{g}(x_{k}) / \tau)$ denote the global model’s smoothed output, and $\hat{y}_{k}^{\tau}=\sigma(\mathcal{F}_{k}(x_{k}) / \tau)$ denote the local model’s smoothed output for client $k$.


The ConDist loss combines four distinct components: foreground class grouping (Section~\ref{sec:fg_grouping}) and conditional probability re-weighting (Section~\ref{sec:cond_prob}), which constitute the core formulation, and two optional enhancements: background class grouping (Section~\ref{sec:bg_grouping}) and foreground filtering (Section~\ref{sec:fg_filter}). The optional techniques primarily improve generalization performance on out-of-federation datasets, as demonstrated in our experimental results.

\subsubsection{Foreground Class Grouping} 
\label{sec:fg_grouping}

The ConDist loss primarily targets optimizing the background classes $B_{k}$, and therefore, the class-specific probabilities associated with the foreground classes $F_{k}$ become secondary in this context. To manage this, we adopt a grouping technique akin to the marginal loss approach, which aggregates the probabilities for the foreground classes. We denote these aggregated probabilities for the global and local models as $\hat{y}_{g, F_{k}}^{\tau}$ and $\hat{y}_{k, F_{k}}^{\tau}$, respectively, as defined in Equation~\ref{eq:fg_group}.

\begin{equation}
  \label{eq:fg_group}
  \hat{y}_{g, F_{k}}^{\tau} = \sum_{i \in F_{k}}(\hat{y}_{g}^{\tau})_{i}
  \quad \mathrm{and} \quad
  \hat{y}_{k, F_{k}}^{\tau} = \sum_{i \in F_{k}}(\hat{y}_{k}^{\tau})_{i}
\end{equation}

This grouping technique produces a new probability distribution that encapsulates an accumulated probability for all classes in $F_{k}$ alongside the probabilities for each background class. However, directly applying this formulation to the loss function introduces conflicts in the optimization target. Specifically, since the probabilities of the labeled organs in $\hat{y}_{g}^{\tau}$ cannot be perfectly aligned with the ground truth $y_{k}$, this discrepancy may lead to ambiguous or even erroneous optimization objectives, ultimately degrading the segmentation performance. This conflict will be addressed later in Section~\ref{sec:cond_prob} through conditional probability techniques.

\subsubsection{Background Class Grouping}
\label{sec:bg_grouping}

In medical image segmentation, lesion regions are often small and vary significantly in size, shape, and location. To reduce the risk of inaccurate lesion predictions from the global model, we group unlabeled organs with their respective tumors. This grouping technique increases tolerance for erroneous lesion prediction by reformulating the lesion segmentation problem to an organ segmentation problem. Let $M_{k}$ denote the number of unlabeled organs in client $k$, and $\mathcal{O}_{k}=\{G_{0}, G_{1}, \dots, G_{M_{k}} \}$ where $G_{0} = \{0\}$ is the set including the background class, and $G_{i}$ is the set including the class index of $i$-th organ and all classes of its anomaly areas for $i = 1, 2, \dots M_{k}$. By accumulating the probabilities of each group $G_{i}$, we can define $\hat{y}_{g, G_{i}}$ and $\hat{y}_{k, G_{i}}$ as follow:

\begin{equation}
  \label{eq:bg_group}
  \hat{y}_{g, G_{i}}^{\tau} = \sum_{j \in G_{i}}(\hat{y}_{g}^{\tau})_{j}
  \quad \mathrm{and} \quad
  \hat{y}_{k, G_{i}}^{\tau} = \sum_{j \in G_{i}}(\hat{y}_{k}^{\tau})_{j}
\end{equation}

Additionally, due to the lack of diagnostic confirmation for the predicted lesion areas, the true nature of these lesions remains uncertain. This grouping technique helps to address the issue of misidentification.

\subsubsection{Conditional Probability} 
\label{sec:cond_prob}
The ConDist loss employs conditional probability to prevent interference between the supervised loss and the ConDist loss. This technique is essential for resolving the conflicts between $y_{k}$ and $\hat{y}^{\tau}_{g}$ discussed in Section \ref{sec:fg_grouping}. By assuming that the given probability is a class in $B_{k}$, we can calculate the conditional probabilities $\hat{y}_{g,\mathcal{O}_{k} | B_{k}}^{\tau}$ and $\hat{y}_{k,\mathcal{O}_{k} | B_{k}}^{\tau}$ using the following equation:

\begin{align}
  \label{eq:cond_prob}
  \hat{y}_{g, \mathcal{O}_{k} | B_{k}}^{\tau} = \left(
    \frac{\hat{y}_{g, G_{0}}^{\tau}}{1-\hat{y}_{g, F_{k}}^{\tau}},
    \frac{\hat{y}_{g, G_{1}}^{\tau}}{1-\hat{y}_{g, F_{k}}^{\tau}},
    \hdots,
    \frac{\hat{y}_{g, G_{M_{k}}}^{\tau}}{1-\hat{y}_{g, F_{k}}^{\tau}}
  \right), \\
  \hat{y}_{k, \mathcal{O}_{k} | B_{k}}^{\tau} = \left(
    \frac{\hat{y}_{k, G_{0}}^{\tau}}{1-\hat{y}_{k, F_{k}}^{\tau}},
    \frac{\hat{y}_{k, G_{1}}^{\tau}}{1-\hat{y}_{k, F_{k}}^{\tau}},
    \hdots,
    \frac{\hat{y}_{k, G_{M_{k}}}^{\tau}}{1-\hat{y}_{k, F_{k}}^{\tau}}
  \right),
\end{align}

This conditional probability formulation ensures that $\hat{y}_{g,\mathcal{O}_{k} | B_{k}}^{\tau}$ and $\hat{y}_{k,\mathcal{O}_{k} | B_{k}}^{\tau}$ are independent of any foreground classes in $F_{k}$. By isolating the background classes in this conditional probability formulation, the ConDist loss can effectively minimize errors related to unlabeled classes without affecting the supervised learning process.

\subsubsection{Foreground Filtering} 
\label{sec:fg_filter}
Foreground filtering is a crucial technique for improving the accuracy of pseudo-labels in distillation learning for unlabeled classes. Since human organs do not overlap, incorrect segmentation of foreground classes $F_{k}$ in the global model’s prediction $\hat{y}^{\tau}_{g}$ can be identified using the partial ground truth labels $y_{k}$. To leverage this characteristic and improve segmentation performance, we introduce a binary mask $\mathcal{M}$ for foreground filtering. This mask shares the spatial shape of the ground truth labels $y_{k}$ and is defined in Equation~\ref{eq:fg_filter} as follows:
\begin{equation}
  \label{eq:fg_filter}
  \mathcal{M}=
  \begin{cases}
    0 & (\mathop{\arg\max}_{i}(\hat{y}_{g}^{\tau})_{i} \in F_{k}) \lor  
        (\mathop{\arg\max}_{i}(y_{k})_{i} \in F_{k}) \\
    1 & \text{Otherwise}
  \end{cases}
\end{equation}
The mask $\mathcal{M}$ is applied via element-wise multiplication to both $\hat{y}_{g,\mathcal{O}_{k}|B_{k}}^{\tau}$ and $\hat{y}_{k,\mathcal{O}_{k}|B_{k}}^{\tau}$ to excludes foreground regions.

The binary mask $\mathcal{M}$ ensures the effectiveness of the distillation process by filtering out irrelevant or incorrectly predicted areas. Specifically, it excludes regions corresponding to labeled classes in $F_{k}$, including correctly segmented, under-segmented, and over-segmented areas. In correctly segmented regions, the probabilities of the unlabeled classes are irrelevant due to the availability of ground truth labels; hence, these regions are filtered out and handled by the supervised loss. For incorrectly segmented areas—whether under-segmented or over-segmented—the mask discards the global model’s erroneous predictions, preventing the local model from learning inaccurate information. Moreover, the mask helps mitigate numerical instability during loss calculation by ensuring the denominator in Equation~\ref{eq:cond_prob} does not approach zero.

\subsubsection{ConDist Loss for Segmentation}
In our proposed conditional distillation, we aim to minimize the distance between $\hat{y}_{k,\mathcal{O}_{k}|B_{k}}^{\tau}$ and $\hat{y}_{g,\mathcal{O}_{k}|B_{k}}^{\tau}$. In the context of segmentation, the soft Dice loss $\mathcal{L}_{Dice}$ is a popular option for knowledge distillation because the Dice score is the primary metric for evaluating segmentation model performance. By combining the foreground filter $\mathcal{M}$ and the soft Dice loss, we define the ConDist loss $\mathcal{L}_{ConDist}$ in Equation~\ref{eq:condist_loss} as follows:

\begin{equation}
  \label{eq:condist_loss}
  \mathcal{L}_{ConDist} =
    \mathcal{L}_{Dice}(
      \mathcal{M} \cdot \hat{y}_{k,\mathcal{O}_{k}|B_{k}},
      \mathcal{M} \cdot \hat{y}_{g,\mathcal{O}_{k}|B_{k}}
    )
\end{equation}

\subsection{Combining Supervised and ConDist Loss}

The total loss function $\mathcal{L}_{total}$, as defined in Equation~\ref{eq:total_loss}, combines the supervised loss $\mathcal{L}_{sup}$ and the ConDist loss $\mathcal{L}_{ConDist}$. A weighting parameter $\lambda$ is introduced to adjust the influence of the ConDist loss during training. This parameter is initialized with a small value and increased linearly throughout training to allow the global model to stabilize before its knowledge is used more heavily in local training.

\begin{equation}
  \label{eq:total_loss}
  \mathcal{L}_{total} = \mathcal{L}_{sup} + \lambda \mathcal{L}_{ConDist}
\end{equation}

By incrementally increasing $\lambda$, we ensure that the model gradually incorporates global knowledge while still prioritizing the available local labels in early FL rounds.

\section{Experiments}
\label{sec:experiments}

In this section, we present our experimental design, which includes details about the datasets, federated learning configurations, model architectures, and evaluation criteria.

\subsection{Datasets and Partial-Label Configuration}
\label{sec:exp_settings}

To simulate realistic cross-institutional conditions, each federated client provides annotations for only a subset of the target structures. For all datasets utilized in federated training, we partition data into training, validation, and test sets according to ratios of 0.6, 0.2, and 0.2, respectively.

\vspace{0.5em}
\noindent
\textbf{Three-dimensional CT Cohort.}
The federated group consists of three organ-specific subsets from the Medical Segmentation Decathlon (MSD)~\cite{Antonelli2021-io}: liver (including tumors), pancreas (including tumors), and spleen. Additionally, kidney and tumor CT data from the KiTS19 dataset~\cite{Heller2021} are included. The AMOS22 dataset~\cite{ji2022amos} is reserved exclusively for out-of-federation evaluation.

\noindent
\textbf{Two-dimensional Chest X-ray Cohort.}
The federated cohort comprises four distinct datasets: Montgomery~\cite{Jaeger2014-ui}, Shenzhen~\cite{Jaeger2014-yo}, JSRT~\cite{Shiraishi2000-ol}, and the SIIM–ACR pneumothorax challenge set~\cite{siim-acr}. Montgomery and Shenzhen datasets include annotations for both lungs, JSRT provides annotations for both lungs and the heart, and SIIM–ACR includes annotations of pneumothorax regions. During training and validation, only the left lung annotations from Montgomery, the right lung annotations from Shenzhen, the heart annotations from JSRT, and the pneumothorax annotations from SIIM–ACR are utilized.

\subsection{Model Architectures and Federated Learning Setup}
\label{sec:fl_setup}

For 3D abdominal CT segmentation, we use MedNeXt-Base~\cite{mednext} with kernel size 3, trained for $120,000$ steps. For 2D chest X-ray segmentation, we adopt DeepLabV3+~\cite{Chen2017-ht} with a ConvNeXtV2-Tiny encoder~\cite{Woo2023ConvNeXtV2}, trained for $30,000$ steps. Both models are optimized using AdamW with cosine annealing schedules and an initial learning rate of $10^{-3}$.

Federated learning experiments are conducted with one central server and four parallel clients. In the 3D experiments, the subsets from the Medical Segmentation Decathlon (MSD) and the KiTS19 dataset are assigned to individual clients. The 2D experiments utilize the Montgomery, Shenzhen, JSRT, and SIIM–ACR datasets, each mapped to a separate client. All training is performed on NVIDIA V100 GPUs using PyTorch~\cite{PyTorch}, NVFLare~\cite{NVFLARE}, and MONAI~\cite{Jorge_Cardoso2022-wb}.

We compare ConDistFL against three standard federated learning optimizers: FedAvg~\cite{fedavg}, FedProx~\cite{fedprox}, and FedOpt~\cite{fedopt}. Additionally, we evaluate its performance against published partial-label methods, including MenuNet~\cite{xu2023federated}, FedIOD~\cite{wan2024fediod}, and the approach by Kim et al.~\cite{Kim2024-ci}, all of which are implemented according to the training schedules and hyperparameters recommended in their respective original papers. The remaining baseline methods adopt the previously described client-side settings.

Unless otherwise specified, FedAvg, FedProx, and FedOpt minimize the marginal loss outlined in Section \ref{sec:sup_loss}. ConDistFL augments this supervised objective with the Conditional Distillation loss, resulting in the combined loss defined in Equation \eqref{eq:total_loss}. For the FedProx method, the proximal coefficient is fixed at $\mu = 10^{-6}$, while FedOpt employs server-side stochastic gradient descent (SGD) with a learning rate of 1.0 and a momentum of 0.6. ConDistFL uses a temperature of $\tau = 0.5$ and progressively increases the distillation weight $\lambda$ from 0.01 to 1.0 throughout the training process.

The federated experiments are conducted over 120 communication rounds for the 3D data, with 1,000 local optimization steps per round, and 30 rounds for the 2D data, consisting of 500 steps per round. For comparative purposes, standalone (non-federated) models are also trained on a single GPU using the same optimization settings as their respective 3D or 2D counterparts.

\section{Results}

This section details ConDistFL's performance through six subsections. Section~\ref{sec:qe} reports baseline Dice scores on federated 3D CT and 2D chest X-ray test sets. Section~\ref{sec:oof} assesses robustness on the external AMOS22 cohort. Section~\ref{sec:compute_time} examines GPU processing time and communication overhead. Section~\ref{sec:dice_dist} analyzes client-level variability by comparing global and local Dice-score distributions. Section~\ref{sec:ablation_oof} conducts an ablation study on foreground filtering and background grouping. Finally, Section~\ref{sec:visualization} visualizes representative segmentation results. Dice scores remain the primary evaluation metric throughout.

\subsection{Quantitative Evaluation}
\label{sec:qe}

\begin{table*}[tb]
\caption{In‑federation test Dice scores (higher is better) for 3D abdominal CT segmentation (top block) and 2D chest X‑ray segmentation (bottom block). Bold values highlight the best metric in each column. An asterisk (*) marks structures that were excluded from training and validation and were used only during testing.}

\label{tab:main_results}

\centering
\begin{tabular}{@{\extracolsep{4pt}}lcccccccc@{}}\\
  \multicolumn{9}{c}{\large{\textbf{3D Abdominal Segmentation}}} \\
  \hline \\ [-1em]
  \, \multirow{2}{*}{Method} & Average
         & \multicolumn{2}{c}{Kidney}   & \multicolumn{2}{c}{Liver}
         & \multicolumn{2}{c}{Pancreas} & Spleen \, \\
         & Dice $\uparrow$
         & Organ & Tumor & Organ & Tumor
         & Oragn & Tumor & Organ \, \\
  \hline \\ [-1em]
  \, Standalone     & 0.8167
                    & \textbf{0.9563} & 0.8116
                    & 0.9520 & 0.7265
                    & 0.7846 & 0.5126
                    & 0.9631 \, \\
  \, FedAvg         & 0.8091
                    & 0.9536 & 0.7766
                    & 0.9608 & 0.7241
                    & 0.7824 & 0.5041
                    & 0.9618 \, \\
  \, FedProx        & 0.7432 
                    & 0.7736 & 0.7481
                    & 0.8973 & 0.5968
                    & 0.7893 & 0.5049
                    & 0.8923 \, \\
  \, FedOpt         & 0.7598
                    & 0.6767 & 0.7751
                    & 0.8915 & 0.6908
                    & 0.7841 & \textbf{0.5435}
                    & 0.9568 \, \\
  \, MenuNet        & 0.6999
                    & 0.9465 & 0.6296
                    & 0.8560 & 0.4846
                    & 0.7607 & 0.2812
                    & 0.9408 \, \\
  \, FedIOD         & 0.6368
                    & 0.9263 & 0.4981
                    & 0.9477 & 0.4681
                    & 0.6719 & 0.0003
                    & 0.9448 \, \\
  \, Kim et al.     & 0.7330 
                    & 0.9430 & 0.6734
                    & 0.9386 & 0.6439
                    & 0.7207 & 0.3778
                    & 0.8334 \, \\
  \, ConDistFL      & \textbf{0.8235} 
                    & 0.9547 & \textbf{0.8247}
                    & \textbf{0.9613} & \textbf{0.7322}
                    & \textbf{0.7975} & 0.5278
                    & \textbf{0.9664} \, \\
  \hline
\end{tabular}

\centering
\begin{tabular}{@{\extracolsep{3pt}}lcccccccccc@{}}\\
\multicolumn{11}{c}{\large{\textbf{2D Chest X-ray Segmentation}}} \\
\hline \\ [-1em]
\, \multirow{2}{*}{Method} & \multicolumn{2}{c}{Average Dice $\uparrow$}
          & \multicolumn{2}{c}{Montgomery} & \multicolumn{2}{c}{Shenzhen} & \multicolumn{3}{c}{JSRT} & \multicolumn{1}{c}{SIIM-ACR} \\

       & \multicolumn{1}{c}{Exclude*} & \multicolumn{1}{c}{Include*}
       & \multicolumn{1}{c}{L. Lung} & \multicolumn{1}{c}{R. Lung*}
       & \multicolumn{1}{c}{L. Lung*} & \multicolumn{1}{c}{R. Lung}
       & \multicolumn{1}{c}{L. Lung*} & \multicolumn{1}{c}{R. Lung*} & \multicolumn{1}{c}{Heart}
       & \multicolumn{1}{c}{Pneumothorax} \\
\hline \\ [-1em]
\, Standalone       & 0.9299 & 0.9456
                 & \textbf{0.9809} & 0.9544
                 & 0.9565 & 0.9699
                 & 0.9626 & 0.9715 & \textbf{0.9544}
                 & 0.8146 \\
\, FedAvg           & 0.9147 & 0.8281 
                 & 0.9731 & 0.6689
                 & 0.5236 & 0.9641
                 & 0.9137 & 0.8601 & 0.9242
                 & 0.7974 \\
\, FedProx          & 0.9280 & 0.7536 
                 & 0.9791 & 0.4367
                 & 0.5032 & 0.9663
                 & 0.8432 & 0.5338 & 0.9426
                 & 0.8240 \\
\, FedOpt           & 0.8354 & 0.7614 
                 & 0.6984 & 0.8238
                 & 0.4837 & 0.9348
                 & 0.6251 & 0.8173 & 0.9052
                 & 0.8033 \\
\, FedIOD           & 0.9072 & 0.9281
                 & 0.9673 & 0.9529
                 & 0.9285 & 0.9608
                 & 0.9499 & 0.9644 & 0.9283
                 & 0.7725 \\
\, ConDistFL     & \textbf{0.9325} & \textbf{0.9489}
                 & 0.9803 & \textbf{0.9674}
                 & \textbf{0.9577} & \textbf{0.9702}
                 & \textbf{0.9630} & \textbf{0.9728} & 0.9537
                 & \textbf{0.8258} \\
\hline
\end{tabular}
\end{table*}

Table \ref{tab:main_results} present a summary of the Dice scores achieved by the best global model derived from each federated method, as well as from non-federated stand-alone models. In the context of the in-federation 3D abdominal CT benchmarks, ConDistFL demonstrates superiority by attaining the highest average Dice score, surpassing all other competitors in the clinically significant categories pertaining to kidney, liver, and spleen, which encompass both organ and tumor segmentation. The disparity in performance is particularly evident in lesion segmentation, where all other federated approaches exhibit a decline in accuracy compared to their stand-alone counterparts.

Furthermore, in the evaluation of the 2D chest X-ray benchmarks, ConDistFL again achieves the highest mean Dice score, leading among the other federated methods in three of the four specified anatomical structures while matching the best competing score for the heart category. It is important to note that Dice scores for the left lung, right lung, and heart categories are calculated exclusively on images with non-empty reference masks, whereas pneumothorax scoring adheres to the SIIM–ACR protocol, which ascribes a Dice score of 1.0 to accurately predicted empty masks and 0.0 to false-positive non-empty masks. Additionally, ConDistFL exhibits the strongest performance on test-only labels, thereby demonstrating effective knowledge transfer to clients that have not encountered those annotations during the training phase.

These findings establish ConDistFL as the most accurate among the federated methods examined, across both volumetric CT and projection X-ray modalities, as well as in relation to both organ and lesion segmentation targets.

\subsection{Generalizability on Out-of-Federation Dataset}
\label{sec:oof}


\begin{figure*}[t]
  \centering
  \includegraphics[width=\textwidth]{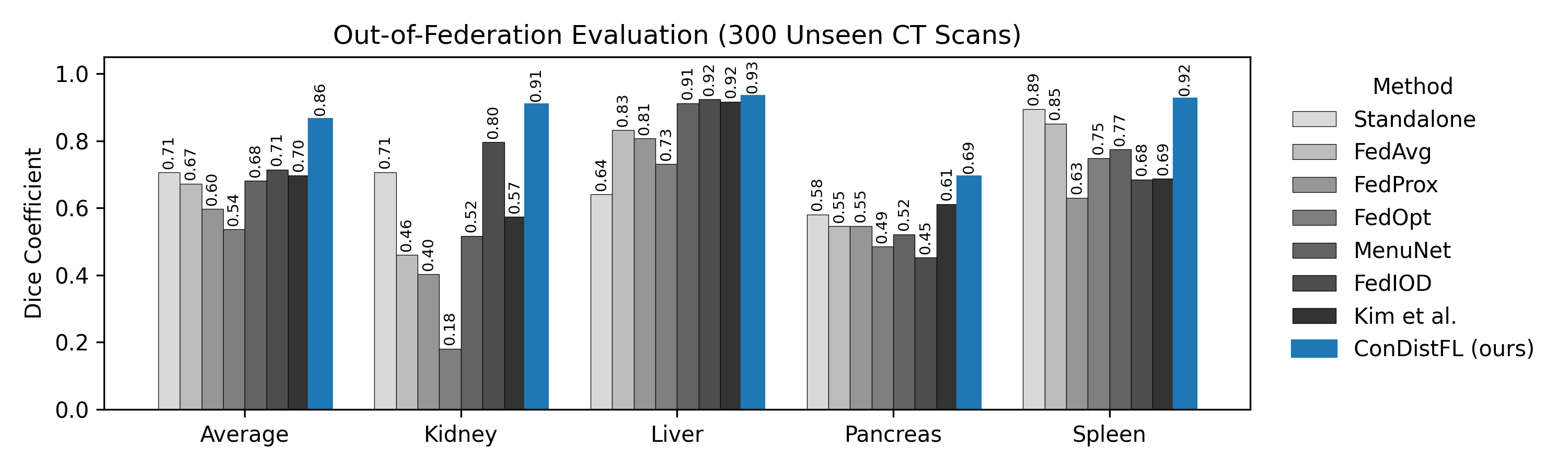}
  \caption{Out‑of‑federation Dice coefficients (AMOS22 dataset, \textit{n}=300 unseen CT scans).  
           ConDistFL is highlighted in blue and achieves the highest average Dice as well as the best class‑specific scores for kidney, liver, pancreas, and spleen.}
  \label{fig:mos_external_test}
\end{figure*}

Figure~\ref{fig:mos_external_test} presents the Dice scores obtained on the AMOS22 dataset, which comprises an external multi-phase CT set that was not utilized during the training phase. The reference masks integrate each organ with its corresponding tumor, and therefore, our evaluation similarly combines organ–lesion predictions (for instance, the kidney along with the kidney tumor). The only exception to this methodology is noted in the results for FedIOD regarding the pancreas, where the tumor channel has been excluded to address the systematic misclassification of background as tumor by this method. This adjustment ensures a more accurate assessment of the pancreas while leaving the evaluations of all other methods unaltered.

ConDistFL achieves the highest mean Dice score and outperforms the competing methods for every organ evaluated, showing no significant decline in performance compared to its in-federation results. In contrast, among the traditional FL baselines, FedAvg demonstrates competitive performance but exhibits lower Dice scores specifically for the kidney and spleen. Additionally, FedProx and FedOpt experience more pronounced drops in performance when faced with domain shifts.

The partial-label methods MenuNet, FedIOD, and the approach presented by Kim~et al. show better generalization compared to traditional baselines. However, each of these methods records at least one organ for which its average Dice score is lower than that of ConDistFL. These results suggest that the proposed conditional distillation strategy significantly enhances cross-domain robustness compared to the evaluated partial-label frameworks.

\subsection{FL Computation and Communication Efficiency}
\label{sec:compute_time}

\begin{table}[ht]
\centering
\caption{Comparison of federated learning methods in terms of FL GPU hours and data traffic for segmentation tasks. Results are presented separately for 3D abdominal CT segmentation and 2D chest X-ray segmentation. Note: Pre-training time for local teacher models in Kim et al. is excluded.}
\label{tab:efficiency}
\begin{tabular}{lcccc} \\
\multicolumn{5}{c}{\textbf{3D Abdominal CT Segmentation}} \\
\hline
Method     & \multicolumn{1}{l}{\#Rounds}
& \multicolumn{1}{l}{Model Size}
& \multicolumn{1}{l}{Traffic Size}
& \multicolumn{1}{l}{GPU Hours} \\
\hline
FedAvg            &   120 &  40.2\,\text{MiB} &  37.7\,\text{GiB} &  372\,\text{GPUh}  \\
FedProx           &   120 &  40.2\,\text{MiB} &  37.7\,\text{GiB} &  380\,\text{GPUh}  \\
FedOpt            &   120 &  40.2\,\text{MiB} &  37.7\,\text{GiB} &  379\,\text{GPUh}  \\
MenuNet           &   400 & 193.9\,\text{MiB} & 581.5\,\text{GiB} & 227\,\text{GPUh} \\
FedIOD            &   500 &  44.5\,\text{MiB} & 162.9\,\text{GiB} & 730\,\text{GPUh} \\
Kim~\emph{et al.} & 1,000 &  60.4\,\text{MiB} & 471.6\,\text{GiB} & 476\,\text{GPUh}  \\
ConDistFL         &   120 &  40.2\,\text{MiB} &  37.7\,\text{GiB} & 410\,\text{GPUh} \\
\hline \\
\multicolumn{5}{c}{\textbf{2D Chest X-ray Segmentation}} \\
\hline
Method     & \multicolumn{1}{l}{\#Rounds}
& \multicolumn{1}{l}{Model Size}
& \multicolumn{1}{l}{Traffic Size}
& \multicolumn{1}{l}{GPU Hours} \\
\hline
FedAvg           &  30 & 336.3\,\text{MiB} &  78.8\,\text{GiB} &  21\,\text{GPUh}  \\
FedProx          &  30 & 336.3\,\text{MiB} &  78.8\,\text{GiB} &  21\,\text{GPUh}  \\
FedOpt           &  30 & 336.3\,\text{MiB} &  78.8\,\text{GiB} &  21\,\text{GPUh}  \\
ConDistFL        &  30 & 336.3\,\text{MiB} &  78.8\,\text{GiB} & 24\,\text{GPUh} \\
\hline
\end{tabular}
\end{table}

Table~\ref{tab:efficiency} summarize the total FL GPU hours and communication traffic for all methods on the 3D and 2D benchmarks. 

For 3D abdominal CT, ConDistFL transmits 37.7\,GiB of parameters, matching the traffic of FedAvg, FedProx, and FedOpt. Its computational cost is 410\,GPU-hours—only a modest increase over these baselines—because the method converges in 120 communication rounds. Partial-label alternatives require substantially more traffic: larger models combined with several hundred rounds raise their totals well beyond ConDistFL’s budget. FedIOD records the highest GPU usage owing to full-volume processing in every epoch, whereas MenuNet attains the lowest GPU hours by validating on a small set of fixed patches rather than the entire validation volume each round.

The 2D chest X-ray experiments exhibit the same trend. ConDistFL shares the 78.8\,GiB traffic of the standard optimizers and adds only a minor GPU-hour overhead attributable to conditional distillation.

Overall, ConDistFL delivers its segmentation improvements while keeping communication demands low and computational requirements practical, demonstrating an effective accuracy–efficiency trade-off across both imaging modalities.

\subsection{Dice Score Distribution Analysis}
\label{sec:dice_dist}

We examine client-level variability by plotting Dice-score distributions for global and local models on both the abdominal CT and chest X-ray benchmarks.

\begin{figure*}
\centering
\includegraphics[width=\textwidth]{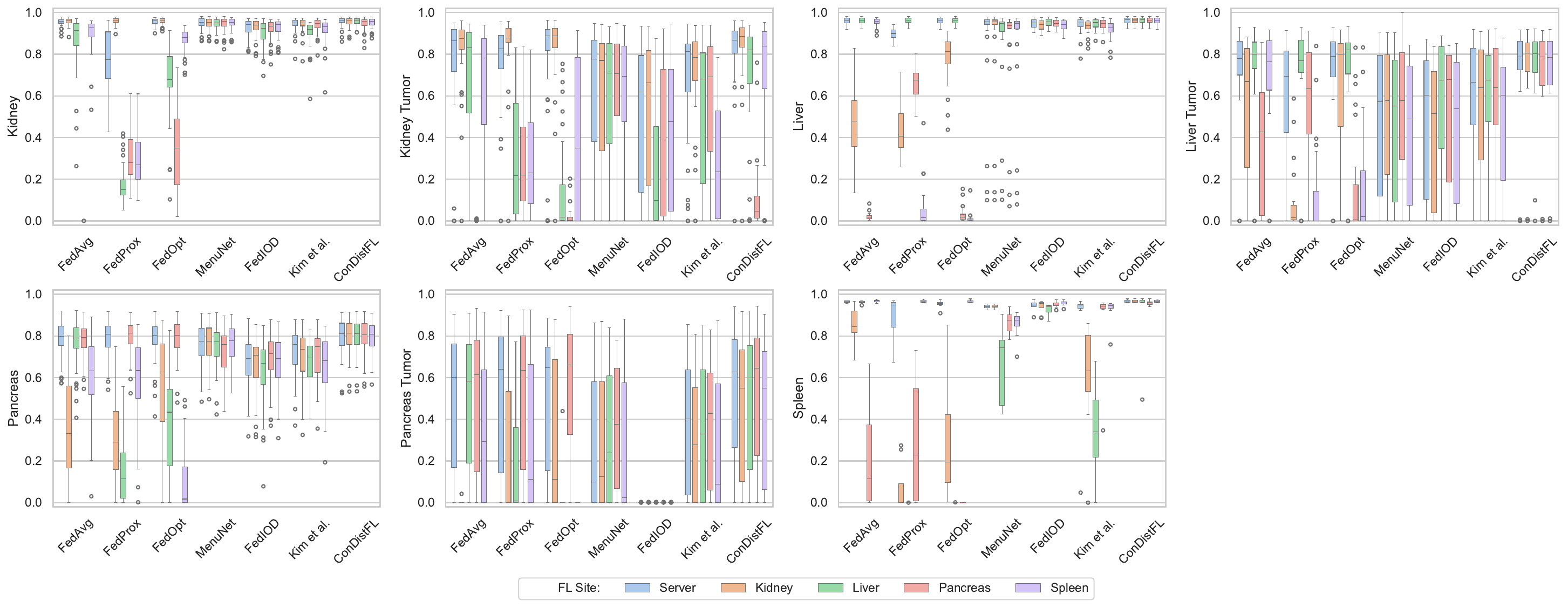}
\caption{Boxplot of test Dice score distributions for global models trained using different federated learning methods on the in-federation test set from the 3D abdominal CT experiments. The Dice scores represent the segmentation performance across multiple clients, with higher scores indicating better agreement between predicted and ground-truth segmentations.}
\label{fig:fl_sites_box_plot}
\end{figure*}

\begin{figure*}
\centering
\includegraphics[width=\textwidth]{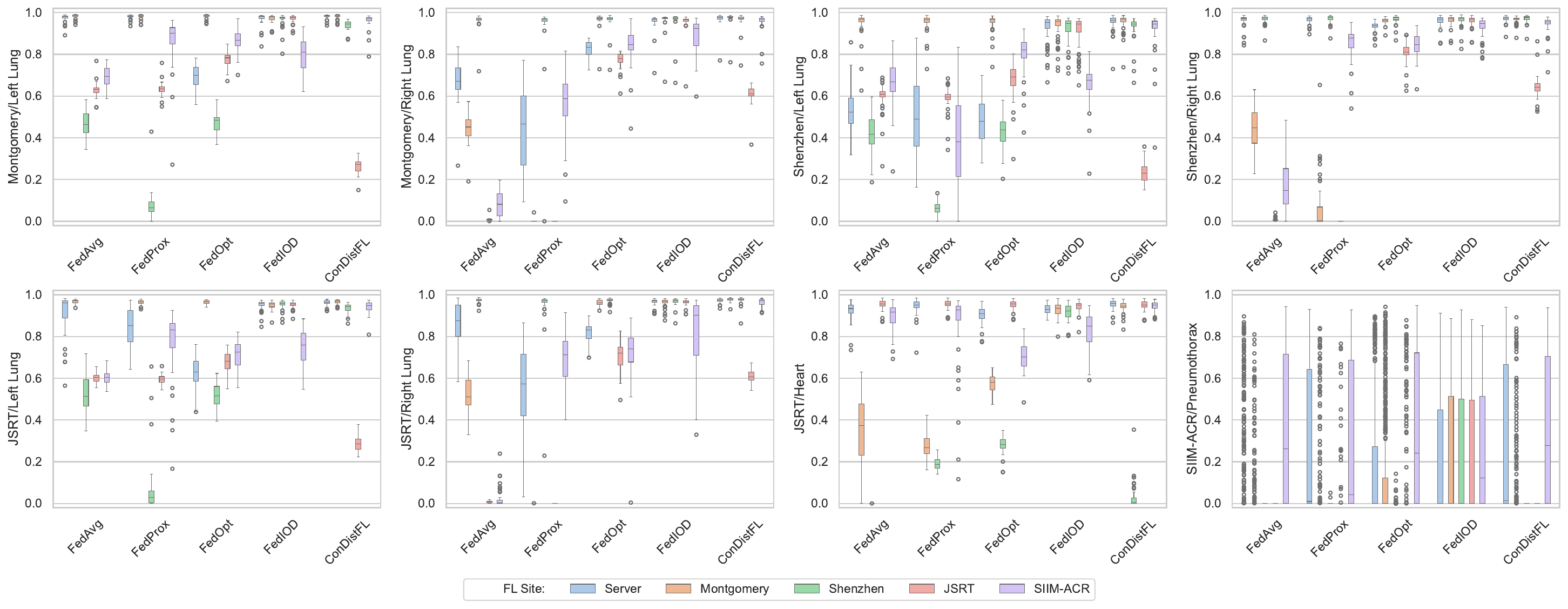}
\caption{Boxplot of test Dice score distributions for global models trained using different federated learning methods on the in-federation test set from the 2D chest X-ray experiments. For the pneumothorax segmentation, the boxplot excludes non-pneumothorax images, displaying only the Dice scores for pneumothorax cases.}
\label{fig:fl_sites_2d}
\end{figure*}

Figures \ref{fig:fl_sites_box_plot} present the client-wise Dice distributions for both global and local models in the abdominal CT benchmark. ConDistFL's local models demonstrate a high degree of alignment with the global model, exhibiting narrow inter-quartile ranges across nearly all organs and tumors. The primary exception is observed in the segmentation of kidney tumors at the pancreas client, where the median Dice score falls below 0.2; nonetheless, it remains superior to competing methods across other classes. In contrast, traditional federated learning approaches, such as FedAvg, FedProx, and FedOpt, display a considerably wider client-to-client variability, with several local models struggling significantly in segmenting liver or pancreas tumors, yielding Dice scores near zero. While MenuNet, FedIOD, and the model developed by Kim \textit{et al.} exhibit improved performance, they still face challenges in maintaining accuracy for tumor classes, with FedIOD notably unable to segment pancreas tumors entirely.

Figure \ref{fig:fl_sites_2d} presents client-wise Dice distributions for the chest-X-ray experiment, comparing global and local models on lung, heart, and pneumothorax labels.  ConDistFL shows markedly tighter inter-quartile ranges than the traditional FL baselines: for the lungs and the heart, it maintains high and uniform Dice scores at nearly every site, even when a label is missing from that client’s training split.  For pneumothorax, the box plot is computed on the 535 images that actually contain the pathology; empty-mask cases are excluded to measure segmentation accuracy rather than prevalence.  Although pneumothorax is the most challenging class, the SIIM-ACR client trained with ConDistFL attains the highest median Dice in the cohort, whereas the baselines show wider dispersion and near-zero medians at several sites.  These observations indicate that conditional distillation improves cross-client consistency and enables the reliable transfer of information to labels that are absent locally, without impairing performance on those already annotated.

\subsection{Ablation Study}
\label{sec:ablation_oof}

\begin{table}[ht]
\centering
\caption{Ablation study on ConDistFL with background class grouping and foreground filtering on the out-of-federation AMOS22 dataset.}
\label{tab:ablation}
\begin{tabular}{ccccccc}\\
\hline \\ [-1em]
Group & Filter & Average & Kidney & Liver & Pancreas & Spleen \\
\hline \\ [-1em]
      &        & 0.8375 & 0.8952 & 0.8687 & \textbf{0.6981} & 0.8881 \\
\checkmark &        & 0.8454 & 0.9063 & 0.9020 & 0.6681 & 0.9052 \\
      & \checkmark  & 0.8361 & 0.8975 & 0.8965 & 0.6496 & 0.9007 \\
\checkmark & \checkmark & \textbf{0.8635} & \textbf{0.9068} & \textbf{0.9318} & 0.6919 & \textbf{0.9235} \\
\hline
\end{tabular}
\end{table}

The ablation study presented in Table \ref{tab:ablation} evaluates two optional components of ConDistFL: background class grouping (Group) and foreground filtering (Filter) within the context of the external AMOS22 cohort. The introduction of background class grouping independently enhances overall accuracy and significantly improves the segmentation of the kidney, liver, and spleen, albeit with a modest decrease in the performance metrics for the pancreas. Similarly, the implementation of foreground filtering in isolation yields a comparable outcome, bolstering the delineation of larger organs while concurrently diminishing the pancreas score, resulting in an overall mean that remains largely unchanged. When both components are utilized, their complementary effects become apparent: the combined configuration attains the highest mean Dice score, further elevates the performance for the kidney, liver, and spleen, and restores the pancreas accuracy to levels approaching the baseline. These findings demonstrate that both grouping and filtering independently contribute to improved large organ segmentation and, when applied collectively, achieve an optimal balance between overall accuracy and organ-specific consistency.

\subsection{Visualization of Segmentation Results}
\label{sec:visualization}
\begin{figure*}[ht]
  \centering
  \includegraphics[width=\textwidth]{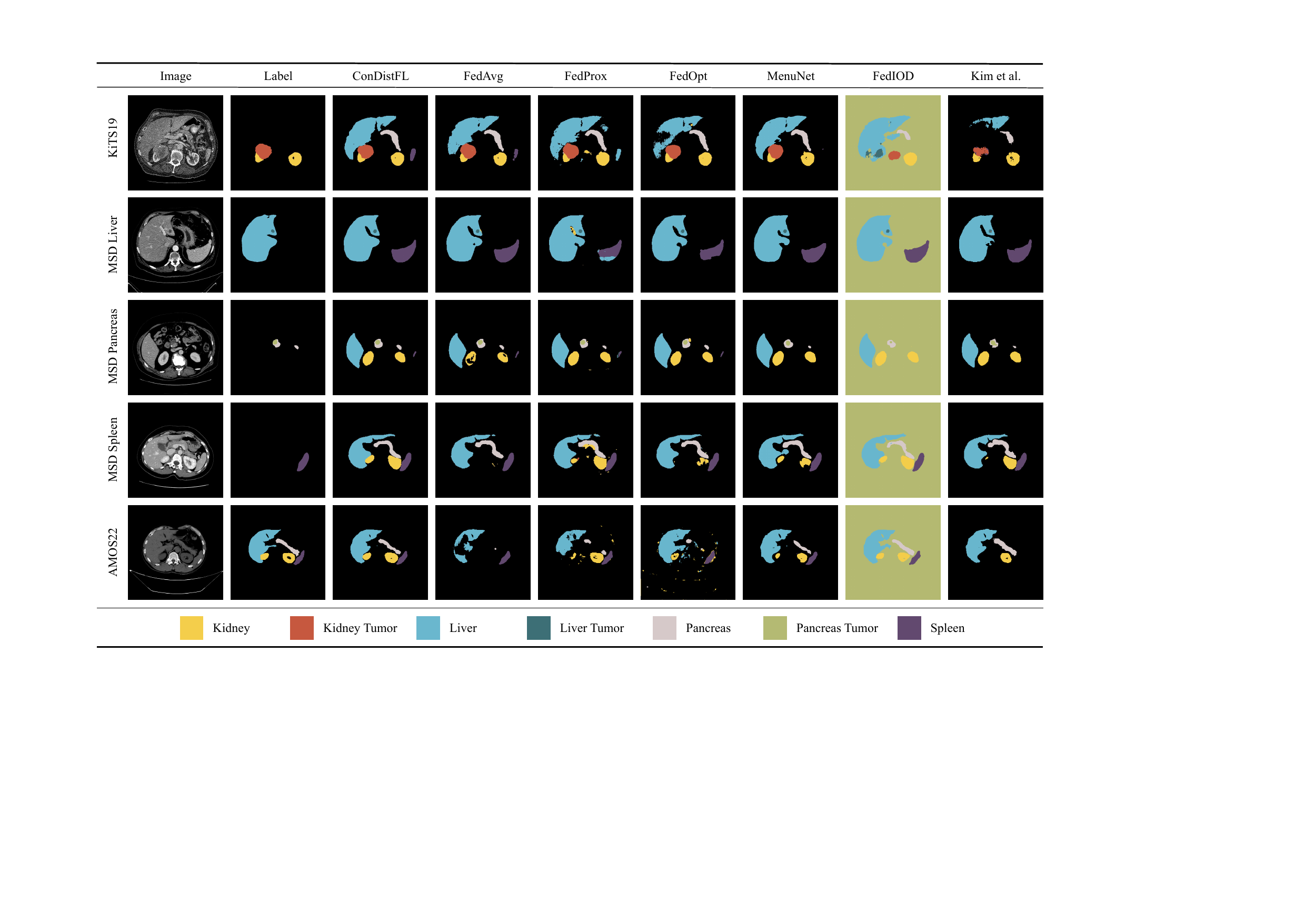}
  \caption{Visual comparison of segmentation results from ConDistFL, FedAvg, FedProx, FedOpt, and Kim et al. on representative slices from the in-federation 3D CT test set. The figure shows the original images, ground truth labels, and predicted segmentations. ConDistFL demonstrates more accurate and consistent segmentation across various organs than other methods.}
  \label{fig:3d_visualization}
\end{figure*}

\begin{figure*}[ht]
  \centering
  \includegraphics[width=\textwidth]{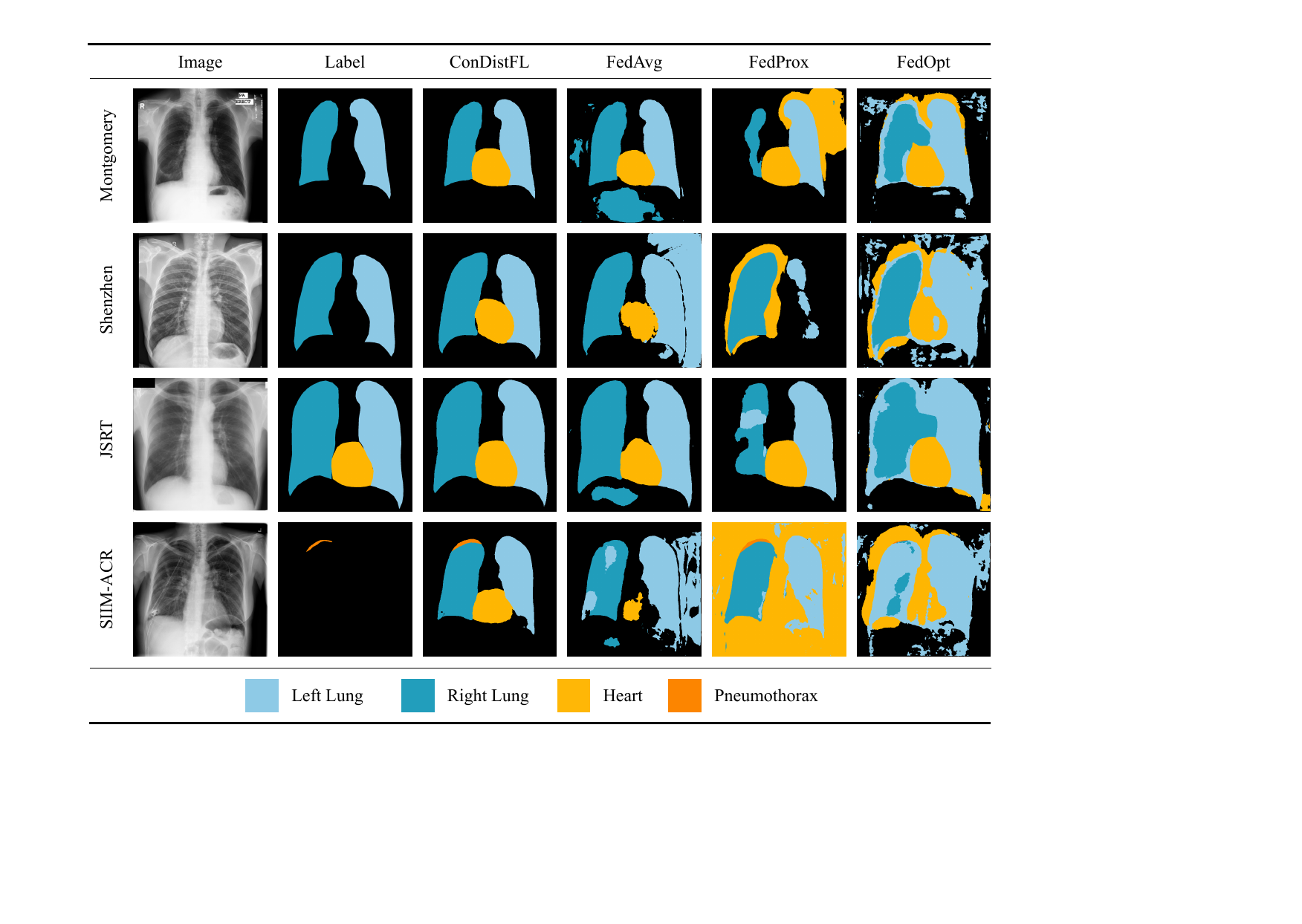}
  \caption{Visual comparison of segmentation results from ConDistFL, FedAvg, FedProx, and FedOpt on representative images from the in-federation 2D chest X-ray test set. The figure shows the original X-ray images, ground truth labels, and predicted segmentations. ConDistFL provides more precise and consistent segmentations for both lung and pneumothorax regions than other methods.}
  \label{fig:2d_visualization}
\end{figure*}

To complement the quantitative evaluation, Figures~\ref{fig:3d_visualization} and~\ref{fig:2d_visualization} present qualitative visualizations of segmentation outcomes for the federated learning global models. These examples include representative slices from the in-federation 3D CT and 2D chest X-ray test sets, as well as an out-of-federation case from the AMOS22 dataset, displaying the original images, ground truth labels, and predictions from ConDistFL and baseline methods.

Across both imaging modalities, ConDistFL produces anatomically accurate and consistent segmentations, particularly in challenging cases involving unlabeled or partially labeled structures. In the 3D CT experiments, ConDistFL demonstrates precise delineation of organs and tumors, even when confronted with images exhibiting lower quality or contrast conditions distinct from those in the training set. This is most evident in the AMOS22 example, where the chosen image is a non-contrast CT scan—a contrast phase entirely absent from the training data. While traditional federated learning methods such as FedAvg and FedOpt frequently yield incomplete or fragmented segmentations under such domain shifts, ConDistFL is the only approach that successfully segments all target organs in the AMOS22 image without substantial errors. This result underscores the robustness of ConDistFL to significant variations in imaging protocol and reinforces its ability to generalize beyond the characteristics of the training domains.

Partial-label baselines, including MenuNet, FedIOD, and the approach of Kim et al., show improvement over traditional methods, but often encounter under-segmentation or misclassification in smaller or less consistently annotated organs, such as the pancreas or kidney tumors. For example, FedIOD is prone to over-segmentation in the pancreas tumor region, frequently misclassifying background as tumor, while other methods may miss entire organs in certain cases.

In the 2D chest X-ray visualizations, ConDistFL again maintains accurate and smooth segmentations for both training-labeled and test-only organs, including the lungs and heart. Its performance is particularly evident in pneumothorax detection, where it produces clear and specific segmentations on the SIIM-ACR dataset, a task where traditional FL baselines typically underperform or fail to produce valid masks. By comparison, baseline models such as FedAvg and FedProx often show over- or under-segmentation, especially for structures absent from local training data, while FedOpt frequently exhibits inconsistent or fragmented predictions.

In summary, these qualitative results support the quantitative findings, demonstrating that ConDistFL achieves greater segmentation consistency and reliability across datasets and label configurations, particularly in challenging or underrepresented scenarios. The visual comparisons reinforce the value of conditional distillation for robust federated learning in multi-site medical image segmentation.

\section{Discussion}

This study presents ConDistFL, a communication-efficient and architecture-agnostic framework for federated medical image segmentation in settings with partially labeled data. Here, we discuss the key findings, implications of our results, practical advantages over existing federated learning methods, and examine the generalizability of our approach in real-world clinical applications.


A central challenge in federated medical image segmentation is the prevalence of partial and heterogeneous label distributions across collaborating institutions. In realistic multicenter datasets, each site may annotate a distinct subset of anatomical structures or pathologies, resulting in non-overlapping label spaces and incomplete supervision for certain classes. Such heterogeneity undermines traditional federated learning approaches, which rely on a consistent label set and uniform supervision across all clients. Our results demonstrate that ConDistFL addresses these limitations by enabling effective cross-site knowledge transfer, even when key labels are absent from individual client datasets. In both 3D abdominal CT and 2D chest X-ray benchmarks, ConDistFL achieves high and consistent Dice scores for both training-annotated and test-only structures, outperforming existing FL methods in the most challenging partially supervised scenarios. In particular, ConDistFL mitigates the severe performance drops observed in traditional FL algorithms on test-only labels, as evidenced by the stable Dice score distributions across clients (see Figures~\ref{fig:fl_sites_box_plot} and~\ref{fig:fl_sites_2d}). These findings underscore the importance of conditional distillation for federated learning, particularly in real-world clinical settings characterized by incomplete and unbalanced label availability.

Importantly, ConDistFL preserves client autonomy by allowing each site to focus training on the structures relevant to its local annotation protocol, while still benefiting from the broader federated knowledge pool. This property enables practical deployment in diverse clinical environments, supporting both cross-organ learning and specialization without requiring costly relabeling or centralized harmonization of label sets. The robustness of ConDistFL to missing or non-overlapping labels provides a significant advantage for large-scale collaborations, where harmonized multi-organ annotation is rarely feasible.


ConDistFL further distinguishes itself from prior methods by achieving robust generalizability to both in-federation and out-of-federation datasets, demonstrating strong segmentation accuracy and stable learning across clients with diverse and partially labeled data. Unlike MenuNet~\cite{xu2023federated} and FedIOD~\cite{wan2024fediod}, ConDistFL does not require custom model layers or manual site-specific adaptations, making it architecture-agnostic and straightforward to integrate into existing clinical workflows. Compared to approaches reliant on pseudo-labeling or local teacher ensembles~\cite{Kim2024-ci}, ConDistFL’s conditional distillation loss provides a more reliable and stable learning signal, reducing the risk of performance collapse on unlabeled or underrepresented structures. Effective generalization to external datasets and heterogeneous client populations is essential for practical medical AI applications, where data distributions and annotation practices frequently differ among institutions. By delivering robust performance without site-specific engineering, ConDistFL offers a scalable pathway for reliable federated deployment in diverse clinical contexts.


Finally, ConDistFL demonstrates strong computational and communication efficiency, with only a modest increase in GPU hours relative to standard federated learning methods and maintaining minimal communication overhead. Detailed computational comparisons provided in Section~\ref{sec:compute_time} confirm that ConDistFL’s efficiency is comparable to or better than baseline approaches, significantly outperforming methods requiring numerous communication rounds or large model updates. These efficiency gains are particularly relevant in multi-center clinical studies and real-world federated deployments, where computational resources and network bandwidth are often limited. By minimizing additional infrastructure burdens while ensuring robust segmentation performance, ConDistFL provides a practically feasible and scalable solution for collaborative medical AI development across diverse clinical environments.

\section{Limitations and Future Directions}

While ConDistFL substantially improves segmentation accuracy and consistency across diverse label distributions, several limitations remain. Despite conditional distillation enhancing cross-client stability, residual performance variability persists, such as difficulty segmenting kidney tumors at pancreas-specialized sites. Additionally, our evaluation was limited to specific datasets and client configurations; broader validation on diverse clinical datasets, encompassing varied imaging protocols and patient demographics, is necessary to confirm further generalizability.

Furthermore, this study employed a fixed aggregation strategy and did not explicitly address scenarios with extreme label sparsity or severe class imbalance. Future research should examine adaptive aggregation techniques sensitive to varying label availability and quality. Extending ConDistFL to semi-supervised or weakly supervised contexts and evaluating integration into practical federated clinical workflows are also promising research directions to enhance clinical translation.

\section{Conclusion}

This work introduces conditional distillation as an effective solution for federated learning on partially labeled medical imaging data. ConDistFL achieves robust and consistent segmentation across diverse clients and out-of-federation cohorts while maintaining efficiency and practicality suitable for real-world deployment. By enabling privacy-preserving collaboration without the need for exhaustive annotation or custom model modifications, ConDistFL advances the scalability and generalizability of multi-site medical AI studies. This approach lays the foundation for more accessible and label-efficient clinical AI, supporting the broader adoption of federated learning in healthcare.

\section*{References}

\bibliographystyle{IEEEtran}
\bibliography{references}

\begin{thebibliography}{10}
\providecommand{\url}[1]{#1}
\csname url@samestyle\endcsname
\providecommand{\newblock}{\relax}
\providecommand{\bibinfo}[2]{#2}
\providecommand{\BIBentrySTDinterwordspacing}{\spaceskip=0pt\relax}
\providecommand{\BIBentryALTinterwordstretchfactor}{4}
\providecommand{\BIBentryALTinterwordspacing}{\spaceskip=\fontdimen2\font plus
\BIBentryALTinterwordstretchfactor\fontdimen3\font minus \fontdimen4\font\relax}
\providecommand{\BIBforeignlanguage}[2]{{%
\expandafter\ifx\csname l@#1\endcsname\relax
\typeout{** WARNING: IEEEtran.bst: No hyphenation pattern has been}%
\typeout{** loaded for the language `#1'. Using the pattern for}%
\typeout{** the default language instead.}%
\else
\language=\csname l@#1\endcsname
\fi
#2}}
\providecommand{\BIBdecl}{\relax}
\BIBdecl

\bibitem{landman2015miccai}
B.~Landman, Z.~Xu, J.~E. Iglesias, M.~Styner, T.~R. Langerak, and A.~Klein, ``Miccai multi-atlas labeling beyond the cranial vault—workshop and challenge,'' in \emph{MICCAI Workshop on Multi-Atlas Labeling Beyond the Cranial Vault}, B.~Landman, Z.~Xu, J.~E. Iglesias, M.~Styner, T.~R. Langerak, and A.~Klein, Eds., 2015, pp. 12--12.

\bibitem{ji2022amos}
Y.~Ji, H.~Bai, J.~Yang, C.~Ge, Y.~Zhu, R.~Zhang, Z.~Li, L.~Zhang, W.~Ma, X.~Wan \emph{et~al.}, ``Amos: A large-scale abdominal multi-organ benchmark for versatile medical image segmentation,'' \emph{arXiv preprint arXiv:2206.08023}, 2022.

\bibitem{Wasserthal2023-ry}
J.~Wasserthal, H.-C. Breit, M.~T. Meyer, M.~Pradella, D.~Hinck, A.~W. Sauter, T.~Heye, D.~T. Boll, J.~Cyriac, S.~Yang, M.~Bach, and M.~Segeroth, ``\BIBforeignlanguage{en}{{TotalSegmentator}: Robust segmentation of 104 anatomic structures in {CT} images},'' \emph{\BIBforeignlanguage{en}{Radiol. Artif. Intell.}}, vol.~5, no.~5, p. e230024, Sep. 2023.

\bibitem{Antonelli2021-io}
M.~Antonelli, A.~Reinke, S.~Bakas, K.~Farahani, {AnnetteKopp-Schneider}, B.~A. Landman, G.~Litjens, B.~Menze, O.~Ronneberger, R.~M. Summers, B.~van Ginneken, M.~Bilello, P.~Bilic, P.~F. Christ, R.~K.~G. Do, M.~J. Gollub, S.~H. Heckers, H.~Huisman, W.~R. Jarnagin, M.~K. McHugo, S.~Napel, J.~S. Goli~Pernicka, K.~Rhode, C.~Tobon-Gomez, E.~Vorontsov, H.~Huisman, J.~A. Meakin, S.~Ourselin, M.~Wiesenfarth, P.~Arbelaez, B.~Bae, S.~Chen, L.~Daza, J.~Feng, B.~He, F.~Isensee, Y.~Ji, F.~Jia, N.~Kim, I.~Kim, D.~Merhof, A.~Pai, B.~Park, M.~Perslev, R.~Rezaiifar, O.~Rippel, I.~Sarasua, W.~Shen, J.~Son, C.~Wachinger, L.~Wang, Y.~Wang, Y.~Xia, D.~Xu, Z.~Xu, Y.~Zheng, A.~L. Simpson, L.~Maier-Hein, and M.~Jorge~Cardoso, ``The medical segmentation decathlon,'' \emph{arXiv [eess.IV]}, Jun. 2021.

\bibitem{Heller2021}
N.~Heller, F.~Isensee, K.~H. Maier-Hein, X.~Hou, C.~Xie, F.~Li, Y.~Nan, G.~Mu, Z.~Lin, M.~Han, G.~Yao, Y.~Gao, Y.~Zhang, Y.~Wang, F.~Hou, J.~Yang, G.~Xiong, J.~Tian, C.~Zhong, J.~Ma, J.~Rickman, J.~Dean, B.~Stai, R.~Tejpaul, M.~Oestreich, P.~Blake, H.~Kaluzniak, S.~Raza, J.~Rosenberg, K.~Moore, E.~Walczak, Z.~Rengel, Z.~Edgerton, R.~Vasdev, M.~Peterson, S.~McSweeney, S.~Peterson, A.~Kalapara, N.~Sathianathen, N.~Papanikolopoulos, and C.~Weight, ``The state of the art in kidney and kidney tumor segmentation in contrast-enhanced ct imaging: Results of the kits19 challenge,'' \emph{Medical Image Analysis}, vol.~67, p. 101821, 2021.

\bibitem{rieke2020future}
N.~Rieke, J.~Hancox, W.~Li, F.~Milletari, H.~R. Roth, S.~Albarqouni, S.~Bakas, M.~N. Galtier, B.~A. Landman, K.~Maier-Hein \emph{et~al.}, ``The future of digital health with federated learning,'' \emph{NPJ digital medicine}, vol.~3, no.~1, pp. 1--7, 2020.

\bibitem{condistFL}
P.~Wang, C.~Shen, W.~Wang, M.~Oda, C.-S. Fuh, K.~Mori, and H.~R. Roth, ``Condistfl: Conditional distillation for federated learning from partially annotated data,'' in \emph{Medical Image Computing and Computer Assisted Intervention -- MICCAI 2023 Workshops}, M.~E. Celebi, M.~S. Salekin, H.~Kim, S.~Albarqouni, C.~Barata, A.~Halpern, P.~Tschandl, M.~Combalia, Y.~Liu, G.~Zamzmi, J.~Levy, H.~Rangwala, A.~Reinke, D.~Wynn, B.~Landman, W.-K. Jeong, Y.~Shen, Z.~Deng, S.~Bakas, X.~Li, C.~Qin, N.~Rieke, H.~R. Roth, and D.~Xu, Eds.\hskip 1em plus 0.5em minus 0.4em\relax Cham: Springer Nature Switzerland, 2023, pp. 311--321.

\bibitem{ronneberger2015u}
O.~Ronneberger, P.~Fischer, and T.~Brox, ``U-net: Convolutional networks for biomedical image segmentation,'' in \emph{Medical image computing and computer-assisted intervention--MICCAI 2015: 18th international conference, Munich, Germany, October 5-9, 2015, proceedings, part III 18}, N.~Navab, J.~Hornegger, W.~M. Wells, and A.~Frangi, Eds.\hskip 1em plus 0.5em minus 0.4em\relax Springer, 2015, pp. 234--241.

\bibitem{vnet2016}
\BIBentryALTinterwordspacing
F.~Milletari, N.~Navab, and S.~Ahmadi, ``V-net: Fully convolutional neural networks for volumetric medical image segmentation,'' in \emph{2016 Fourth International Conference on 3D Vision (3DV)}, S.~Savarese, N.~Snavely, D.~Cremers, I.~Reid, and S.~Fidler, Eds.\hskip 1em plus 0.5em minus 0.4em\relax Los Alamitos, CA, USA: IEEE Computer Society, oct 2016, pp. 565--571. [Online]. Available: \url{https://doi.ieeecomputersociety.org/10.1109/3DV.2016.79}
\BIBentrySTDinterwordspacing

\bibitem{Zhou2018-ji}
Z.~Zhou, M.~M.~R. Siddiquee, N.~Tajbakhsh, and J.~Liang, ``{UNet++}: A nested {U}-net architecture for medical image segmentation,'' \emph{arXiv [cs.CV]}, 18~Jul. 2018.

\bibitem{isensee2021nnu}
F.~Isensee, P.~F. Jaeger, S.~A. Kohl, J.~Petersen, and K.~H. Maier-Hein, ``nnu-net: a self-configuring method for deep learning-based biomedical image segmentation,'' \emph{Nature methods}, vol.~18, no.~2, pp. 203--211, 2021.

\bibitem{hatamizadeh2022unetr}
A.~Hatamizadeh, Y.~Tang, V.~Nath, D.~Yang, A.~Myronenko, B.~Landman, H.~R. Roth, and D.~Xu, ``Unetr: Transformers for 3d medical image segmentation,'' in \emph{Proceedings of the IEEE/CVF winter conference on applications of computer vision}, G.~Medioni, K.~W. Bowyer, W.~J. Scheirer, R.~Farrell, C.~Zhao, S.~Anand, and R.~Souvenir, Eds., 2022, pp. 574--584.

\bibitem{hatamizadeh2021swin}
A.~Hatamizadeh, V.~Nath, Y.~Tang, D.~Yang, H.~R. Roth, and D.~Xu, ``Swin unetr: Swin transformers for semantic segmentation of brain tumors in mri images,'' in \emph{International MICCAI brainlesion workshop}, A.~Crimi and S.~Bakas, Eds.\hskip 1em plus 0.5em minus 0.4em\relax Springer, 2021, pp. 272--284.

\bibitem{roy2023mednext}
S.~Roy, G.~Koehler, C.~Ulrich, M.~Baumgartner, J.~Petersen, F.~Isensee, P.~F. Jaeger, and K.~H. Maier-Hein, ``Mednext: transformer-driven scaling of convnets for medical image segmentation,'' in \emph{International Conference on Medical Image Computing and Computer-Assisted Intervention}, H.~Greenspan, A.~Madabhushi, P.~Mousavi, S.~Salcudean, J.~Duncan, T.~Syeda-Mahmood, and R.~Taylor, Eds.\hskip 1em plus 0.5em minus 0.4em\relax Springer, 2023, pp. 405--415.

\bibitem{Isensee2024-fr}
F.~Isensee, T.~Wald, C.~Ulrich, M.~Baumgartner, S.~Roy, K.~Maier-Hein, and P.~F. Jaeger, ``{NnU}-net revisited: A call for rigorous validation in {3D} medical image segmentation,'' \emph{arXiv [cs.CV]}, 15~Apr. 2024.

\bibitem{SHI2021101979}
\BIBentryALTinterwordspacing
G.~Shi, L.~Xiao, Y.~Chen, and S.~K. Zhou, ``Marginal loss and exclusion loss for partially supervised multi-organ segmentation,'' \emph{Medical Image Analysis}, vol.~70, p. 101979, 2021. [Online]. Available: \url{https://www.sciencedirect.com/science/article/pii/S1361841521000256}
\BIBentrySTDinterwordspacing

\bibitem{fang2020multi}
X.~Fang and P.~Yan, ``Multi-organ segmentation over partially labeled datasets with multi-scale feature abstraction,'' \emph{IEEE Transactions on Medical Imaging}, vol.~39, no.~11, pp. 3619--3629, 2020.

\bibitem{zhang2021dodnet}
J.~Zhang, Y.~Xie, Y.~Xia, and C.~Shen, ``Dodnet: Learning to segment multi-organ and tumors from multiple partially labeled datasets,'' in \emph{Proceedings of the IEEE conference on computer vision and pattern recognition}, M.~S. Brown, R.~Sukthankar, T.~Tan, L.~Zelnik, D.~Forsyth, G.~Gkioxari, T.~Tuytelaars, R.~Yang, and J.~Yu, Eds., 2021.

\bibitem{xie2023learning}
Y.~Xie, J.~Zhang, Y.~Xia, and C.~Shen, ``Learning from partially labeled data for multi-organ and tumor segmentation,'' \emph{IEEE Transactions on Pattern Analysis and Machine Intelligence}, 2023.

\bibitem{liu2024cosst}
H.~Liu, Z.~Xu, R.~Gao, H.~Li, J.~Wang, G.~Chabin, I.~Oguz, and S.~Grbic, ``Cosst: Multi-organ segmentation with partially labeled datasets using comprehensive supervisions and self-training,'' \emph{IEEE Transactions on Medical Imaging}, 2024.

\bibitem{liu2023clip}
J.~Liu, Y.~Zhang, J.-N. Chen, J.~Xiao, Y.~Lu, B.~A~Landman, Y.~Yuan, A.~Yuille, Y.~Tang, and Z.~Zhou, ``Clip-driven universal model for organ segmentation and tumor detection,'' in \emph{Proceedings of the IEEE/CVF International Conference on Computer Vision}, J.~Kosecka, J.~Ponce, C.~Schmid, A.~Zisserman, L.~Agapito, Y.~Furukawa, K.~Grauman, K.~He, and I.~Laptev, Eds., 2023, pp. 21\,152--21\,164.

\bibitem{Liu2022-ec}
P.~Liu, X.~Wang, M.~Fan, H.~Pan, M.~Yin, X.~Zhu, D.~Du, X.~Zhao, L.~Xiao, L.~Ding, X.~Wu, and S.~K. Zhou, ``Learning incrementally to segment multiple organs in a {CT} image,'' in \emph{Medical Image Computing and Computer Assisted Intervention -- MICCAI 2022}, L.~Wang, Q.~Dou, P.~T. Fletcher, S.~Speidel, and S.~Li, Eds.\hskip 1em plus 0.5em minus 0.4em\relax Springer Nature Switzerland, 2022, pp. 714--724.

\bibitem{Ji2023-vw}
Z.~Ji, D.~Guo, P.~Wang, K.~Yan, L.~Lu, M.~Xu, J.~Zhou, Q.~Wang, J.~Ge, M.~Gao, X.~Ye, and D.~Jin, ``Continual segment: Towards a single, unified and accessible continual segmentation model of 143 whole-body organs in {CT} scans,'' \emph{arXiv [cs.CV]}, 1~Feb. 2023.

\bibitem{shen2022joint}
C.~Shen, P.~Wang, D.~Yang, D.~Xu, M.~Oda, P.-T. Chen, K.-L. Liu, W.-C. Liao, C.-S. Fuh, K.~Mori \emph{et~al.}, ``Joint multi organ and tumor segmentation from partial labels using federated learning,'' in \emph{International Workshop on Distributed, Collaborative, and Federated Learning}, S.~Albarqouni, S.~Bano, B.~Khanal, X.~Li, I.~Rekik, H.~Roth, D.~Xu, S.~Bakas, M.~J. Cardoso, B.~Landman, C.~Qin, N.~Rieke, and D.~Sheet, Eds.\hskip 1em plus 0.5em minus 0.4em\relax Springer, 2022, pp. 58--67.

\bibitem{xu2023federated}
X.~Xu, H.~H. Deng, J.~Gateno, and P.~Yan, ``Federated multi-organ segmentation with inconsistent labels,'' \emph{IEEE transactions on medical imaging}, vol.~42, no.~10, pp. 2948--2960, 2023.

\bibitem{wan2024fediod}
Q.~Wan, Z.~Yan, and L.~Yu, ``Fediod: Federated multi-organ segmentation from partial labels by exploring inter-organ dependency,'' \emph{IEEE Journal of Biomedical and Health Informatics}, 2024.

\bibitem{Kim2024-ci}
S.~Kim, H.~Park, M.~Kang, K.~H. Jin, E.~Adeli, K.~M. Pohl, and S.~H. Park, ``\BIBforeignlanguage{en}{Federated learning with knowledge distillation for multi-organ segmentation with partially labeled datasets},'' \emph{\BIBforeignlanguage{en}{Med. Image Anal.}}, vol.~95, no. 103156, p. 103156, 1~Jul. 2024.

\bibitem{Jaeger2014-ui}
S.~Jaeger, A.~Karargyris, S.~Candemir, L.~Folio, J.~Siegelman, F.~Callaghan, {Zhiyun Xue}, K.~Palaniappan, R.~K. Singh, S.~Antani, G.~Thoma, {Yi-Xiang Wang}, {Pu-Xuan Lu}, and C.~J. McDonald, ``\BIBforeignlanguage{en}{Automatic tuberculosis screening using chest radiographs},'' \emph{\BIBforeignlanguage{en}{IEEE Trans. Med. Imaging}}, vol.~33, no.~2, pp. 233--245, Feb. 2014.

\bibitem{Jaeger2014-yo}
S.~Jaeger, S.~Candemir, S.~Antani, Y.-X.~J. Wáng, P.-X. Lu, and G.~Thoma, ``\BIBforeignlanguage{en}{Two public chest {X}-ray datasets for computer-aided screening of pulmonary diseases},'' \emph{\BIBforeignlanguage{en}{Quant. Imaging Med. Surg.}}, vol.~4, no.~6, pp. 475--477, Dec. 2014.

\bibitem{Shiraishi2000-ol}
J.~Shiraishi, S.~Katsuragawa, J.~Ikezoe, T.~Matsumoto, T.~Kobayashi, K.~Komatsu, M.~Matsui, H.~Fujita, Y.~Kodera, and K.~Doi, ``\BIBforeignlanguage{en}{Development of a digital image database for chest radiographs with and without a lung nodule: receiver operating characteristic analysis of radiologists' detection of pulmonary nodules: Receiver operating characteristic analysis of radiologists' detection of pulmonary nodules},'' \emph{\BIBforeignlanguage{en}{AJR Am. J. Roentgenol.}}, vol. 174, no.~1, pp. 71--74, Jan. 2000.

\bibitem{siim-acr}
A.~Zawacki, C.~Wn, G.~Shih, J.~Elliott, M.~Fomitchev, M.~Hussain, ParasLakhani, P.~Culliton, and S.~Bao, ``Siim-acr pneumothorax segmentation.'' \emph{Kaggle.}, 2019.

\bibitem{mednext}
S.~Roy, G.~Koehler, C.~Ulrich, M.~Baumgartner, J.~Petersen, F.~Isensee, P.~F. J{\"a}ger, and K.~H. Maier-Hein, ``Mednext: Transformer-driven scaling of convnets for medical image segmentation,'' in \emph{Medical Image Computing and Computer Assisted Intervention -- MICCAI 2023}, H.~Greenspan, A.~Madabhushi, P.~Mousavi, S.~Salcudean, J.~Duncan, T.~Syeda-Mahmood, and R.~Taylor, Eds.\hskip 1em plus 0.5em minus 0.4em\relax Cham: Springer Nature Switzerland, 2023, pp. 405--415.

\bibitem{Chen2017-ht}
L.-C. Chen, G.~Papandreou, F.~Schroff, and H.~Adam, ``Rethinking atrous convolution for semantic image segmentation,'' \emph{arXiv [cs.CV]}, 2017.

\bibitem{Woo2023ConvNeXtV2}
S.~Woo, S.~Debnath, R.~Hu, X.~Chen, Z.~Liu, I.~S. Kweon, and S.~Xie, ``Convnext v2: Co-designing and scaling convnets with masked autoencoders,'' \emph{arXiv preprint arXiv:2301.00808}, 2023.

\bibitem{PyTorch}
\BIBentryALTinterwordspacing
J.~Ansel, E.~Yang, H.~He, N.~Gimelshein, A.~Jain, M.~Voznesensky, B.~Bao, P.~Bell, D.~Berard, E.~Burovski, G.~Chauhan, A.~Chourdia, W.~Constable, A.~Desmaison, Z.~DeVito, E.~Ellison, W.~Feng, J.~Gong, M.~Gschwind, B.~Hirsh, S.~Huang, K.~Kalambarkar, L.~Kirsch, M.~Lazos, M.~Lezcano, Y.~Liang, J.~Liang, Y.~Lu, C.~Luk, B.~Maher, Y.~Pan, C.~Puhrsch, M.~Reso, M.~Saroufim, M.~Y. Siraichi, H.~Suk, M.~Suo, P.~Tillet, E.~Wang, X.~Wang, W.~Wen, S.~Zhang, X.~Zhao, K.~Zhou, R.~Zou, A.~Mathews, G.~Chanan, P.~Wu, and S.~Chintala, ``Pytorch 2: Faster machine learning through dynamic python bytecode transformation and graph compilation,'' in \emph{29th ACM International Conference on Architectural Support for Programming Languages and Operating Systems, Volume 2 (ASPLOS '24)}, N.~Abu-Ghazaleh, R.~Gupta, M.~Musuvathi, and D.~Tsafrir, Eds.\hskip 1em plus 0.5em minus 0.4em\relax ACM, Apr. 2024. [Online]. Available: \url{https://pytorch.org/assets/pytorch2-2.pdf}
\BIBentrySTDinterwordspacing

\bibitem{NVFLARE}
H.~R. Roth, Y.~Cheng, Y.~Wen, I.~Yang, Z.~Xu, Y.-T. Hsieh, K.~Kersten, A.~Harouni, C.~Zhao, K.~Lu, Z.~Zhang, W.~Li, A.~Myronenko, D.~Yang, S.~Yang, N.~Rieke, A.~Quraini, C.~Chen, D.~Xu, N.~Ma, P.~Dogra, M.~Flores, and A.~Feng, ``Nvidia flare: Federated learning from simulation to real-world,'' \emph{IEEE Data Eng. Bull., Vol. 46, No. 1}, Mar. 2023.

\bibitem{Jorge_Cardoso2022-wb}
M.~Jorge~Cardoso, W.~Li, R.~Brown, N.~Ma, E.~Kerfoot, Y.~Wang, B.~Murrey, A.~Myronenko, C.~Zhao, D.~Yang, V.~Nath, Y.~He, Z.~Xu, A.~Hatamizadeh, A.~Myronenko, W.~Zhu, Y.~Liu, M.~Zheng, Y.~Tang, I.~Yang, M.~Zephyr, B.~Hashemian, S.~Alle, M.~Z. Darestani, C.~Budd, M.~Modat, T.~Vercauteren, G.~Wang, Y.~Li, Y.~Hu, Y.~Fu, B.~Gorman, H.~Johnson, B.~Genereaux, B.~S. Erdal, V.~Gupta, A.~Diaz-Pinto, A.~Dourson, L.~Maier-Hein, P.~F. Jaeger, M.~Baumgartner, J.~Kalpathy-Cramer, M.~Flores, J.~Kirby, L.~A.~D. Cooper, H.~R. Roth, D.~Xu, D.~Bericat, R.~Floca, S.~Kevin~Zhou, H.~Shuaib, K.~Farahani, K.~H. Maier-Hein, S.~Aylward, P.~Dogra, S.~Ourselin, and A.~Feng, ``{MONAI}: An open-source framework for deep learning in healthcare,'' \emph{arXiv [cs.LG]}, Nov. 2022.

\bibitem{fedavg}
B.~McMahan, E.~Moore, D.~Ramage, S.~Hampson, and B.~A. y~Arcas, ``Communication-efficient learning of deep networks from decentralized data,'' in \emph{Artificial intelligence and statistics}, A.~Singh and X.~J. Zhu, Eds.\hskip 1em plus 0.5em minus 0.4em\relax PMLR, 2017, pp. 1273--1282.

\bibitem{fedprox}
T.~Li, A.~K. Sahu, M.~Zaheer, M.~Sanjabi, A.~Talwalkar, and V.~Smith, ``Federated optimization in heterogeneous networks,'' \emph{Proceedings of Machine learning and systems}, vol.~2, pp. 429--450, 2020.

\bibitem{fedopt}
S.~Reddi, Z.~Charles, M.~Zaheer, Z.~Garrett, K.~Rush, J.~Kone{\v{c}}n{\`y}, S.~Kumar, and H.~B. McMahan, ``Adaptive federated optimization,'' \emph{arXiv preprint arXiv:2003.00295}, 2020.

\end{thebibliography}

\end{document}